\documentclass[conference]{IEEEtran}
\IEEEoverridecommandlockouts
\usepackage{cite}
\usepackage{amsmath,amssymb,amsfonts}
\usepackage[linesnumbered,ruled,vlined]{algorithm2e}
\usepackage{algpseudocode}
\usepackage{graphicx}
\usepackage{textcomp}
\usepackage{xcolor}
\usepackage{mymacros}
\usepackage{hyperref}

\SetKwInput{KwInput}{Input}                
\SetKwInput{KwOutput}{Output}              
\SetKw{Break}{break}

\begin{document}

\title{Robust Counterexample-guided Optimization for Planning from Differentiable Temporal Logic \\
\thanks{C. Dawson is supported by the NSF GRFP under Grant No. 1745302. The Defense Science and Technology Agency in Singapore provided funds to assist the authors with their research, but this article solely reflects the opinions and conclusions of its authors and not DSTA Singapore or the Singapore Government.}
}

\author{\IEEEauthorblockN{Charles Dawson, Chuchu Fan}
\IEEEauthorblockA{\textit{Department of Aeronautics and Astronautics} \\
\textit{Massachusetts Institute of Technology}\\
Cambridge, USA \\
\texttt{\{cbd, chuchu\}@mit.edu}}
}

\maketitle

\begin{abstract}
Signal temporal logic (STL) provides a powerful, flexible framework for specifying complex autonomy tasks; however, existing methods for planning based on STL specifications have difficulty scaling to long-horizon tasks and are not robust to external disturbances. In this paper, we present an algorithm for finding robust plans that satisfy STL specifications. Our method alternates between local optimization and local falsification, using automatically differentiable temporal logic to iteratively optimize its plan in response to counterexamples found during the falsification process. We benchmark our counterexample-guided planning method against state-of-the-art planning methods on two long-horizon satellite rendezvous missions, showing that our method finds high-quality plans that satisfy STL specifications despite adversarial disturbances. We find that our method consistently finds plans that are robust to adversarial disturbances and requires less than half the time of competing methods. We provide an implementation of our  planner at \url{https://github.com/MIT-REALM/architect}.
\end{abstract}

\begin{IEEEkeywords}
formal methods, differentiable programming, temporal logic
\end{IEEEkeywords}

\section{Introduction \& Related Work}

There is a substantial gap between how many users dream of interacting with intelligent robots and how those robots are programmed in reality. The dream is for the human user to instruct their robot in something not too far from natural language, e.g. ``please visit both the gas station and the grocery store, and make sure you get back here within 30 minutes'', or ``land at one of three landing pads, but stay clear of other aircraft''. Unfortunately, robots today usually expect much more concrete guidance, such as a specific trajectory or feedback controller. As the tasks we wish to assign our robots grow increasingly complex, there is a correspondingly increased need for flexible specification of robot programs and tools to automatically derive concrete plans from those specifications. Moreover, since the real world is unavoidably messy, any plan thus derived must also be robust to unforeseen variation in the environment; the robot must be able to accomplish its plan even when the environment changes.

Luckily, when it comes to flexibly specifying complex tasks, we have a convenient tool in the form of \textit{temporal logic}. There are many flavors of temporal logic, but most relevant to many robotics problems is \textit{signal temporal logic} (or STL), which provides a flexible language for specifying requirements for continuous real-valued signals~\cite{donze10,Sun2022,Pant2017}. STL allows a user to specify a wide range of planning problems by combining logical and temporal operators to express requirements about ordering and dependencies between subtasks. In addition, although the formal syntax of STL can seem opaque at first, it is often quite easy to translate STL formulae into readily-understood natural language. Due to its flexibility, STL is a common choice for specifying robotics problems such as trajectory planning~\cite{Pant2018,Pantazides2022} and combined task and motion planning~\cite{Plaku2016,Sun2022,Takano2021}.

A number of classical methods exist for planning from STL specifications, the most common being abstraction-based methods~\cite{Plaku2016}, mixed-integer optimization-based methods~\cite{Sun2022,yang20}, and nonlinear optimization methods~\cite{Pant2018, Pantazides2022, Leung2020} (other approaches include sampling-based methods such as~\cite{kantaros20} and~\cite{vasile17}). Abstraction-based methods have the longest history; these methods first construct a discrete abstraction (in the form of a graph or automaton) of the continuous state space, then plan over this discrete abstraction~\cite{Plaku2016}. The drawback of abstraction-based methods is that the size of the discrete abstraction grows exponentially with the dimension of the state space, limiting the scalability of these methods.

Other methods, based on mixed-integer optimization, exploit the fact that STL specifications can be expressed as linear constraints with integer variables, and the resulting optimization formulation provide soundness and completeness guarantees. Unfortunately, although mixed-integer optimization is sound and complete, these mixed-integer programs quickly become intractable as the planning horizon increases~\cite{raman15,sadraddini15,yang20}. Some works reduce the size of the program by using timed waypoints instead of a receding horizon~\cite{Sun2022}, but this requires assumptions (such as access to a bounded-error tracking controller) that can be restrictive.

A more recent line of work has focused on solving STL planning problems using nonlinear optimization~\cite{Pant2017,Pant2018,Pantazides2022,Leung2020,Takano2021}. In these approaches, the STL specification is replaced with a continuously differentiable approximation and optimized using local gradient-based methods. These approaches achieve increased generality and scalability by sacrificing completeness and optimality guarantees.

A significant gap in the state of the art is that existing optimization-based STL planners~\cite{Pant2017,Pant2018,Takano2021,Leung2020,Pantazides2022,Sun2022} do not explicitly consider the effects of environmental disturbances while planning. These approaches include some amount of robustness implicitly, typically by maximizing the margin by which a plan satisfies the STL specification, but this is often not sufficient in practice to prevent the plan from failing in response to small changes in the environment. Some methods do explicitly consider robustness to disturbances~\cite{raman15}, but our experiments show that they yield mixed-integer problems that are intractable in practice.

In this paper, we fill this gap by developing a robust planner that uses counterexamples (examples of environmental changes that cause the plan to fail) to refine its plan using nonlinear optimization. This planner relies on an iterative optimization process, inspired by solution methods for multi-player games, that alternates between finding a plan that performs well for all counterexamples seen so far and finding new counterexamples to guide the optimization process. Our framework relies on differentiable simulation and differentiable temporal logic to derive gradients of the plan's performance with respect to both the planning parameters and the environmental disturbance, enabling an efficient search for new plans and counterexamples.

We compare our approach against state-of-the-art methods, including both mixed-integer methods~\cite{raman15} and nonlinear optimization methods~\cite{Pant2018,Pantazides2022,Leung2020}. We find that our method not only finds plans that succeed despite worst-case disturbances from the environment, but it also requires less than half the time of the next-most-successful method. Our approach easily scales to handle long-horizon tasks with complex STL specifications that are not tractable for mixed-integer programming, and the plans found using our method are consistently more robust than those found using existing methods.

\section{Preliminaries}\label{preliminaries}

We begin by introducing the syntax and semantics of signal temporal logic, or STL. STL defines properties about real-valued functions of time $s: \R^+ \mapsto \R^n$ called \textit{signals}. For our purposes, a signal is defined by a finite number of sampled points $(t_i, x_i)$, and we assume that the signal is piecewise-affine in between sampled points and constant after the last sample. Syntactically, an STL formula is constructed from predicates based on functions $\mu: \R^n \mapsto \R$, logical connectives, and temporal operators~\cite{donze13}. The syntax of an STL formula $\psi$ is defined inductively as:
\begin{align*}
    \psi = \text{true}\ |\ \mu(x) \geq 0\ |\ \neg \psi\ |\ \psi_1 \wedge \psi_2\ |\ \psi_1 \ \until_I\ \psi_2
\end{align*}
where $I$ is a closed (but potentially unbounded) time interval and $\until_I$ is the ``until'' operator (read as: within interval $I$, $\psi_1$ must be true until $\psi_2$ becomes true). For convenience, when $I$ is omitted it is assumed to be $[0, \infty)$. Additional temporal operators such as eventually $\eventually_I\ \psi = \text{true } \until_I\ \psi$ and always $\always_I = \neg \eventually_I\ \neg \psi$ follow from this basic syntax, as do logical operators such as $\vee$ and $\implies$.

For any signal $s$, an STL formula is satisfied at a given time $t$ according to the following Boolean semantics~\cite{donze13}:
\begin{alignat*}{3}
    &s, t &&\models \text{true} && \\
    &s, t &&\models \mu(x) \geq 0\quad &&\text{iff}\ \mu(s(t)) \geq 0 \\
    &s, t &&\models \neg \psi &&\text{iff}\ s, t \not\models \psi \\
    &s, t &&\models \psi_1 \wedge \psi_2 &&\text{iff}\ s, t \models \psi_1 \text{ and } s, t \models \psi_2 \\
    &s, t &&\models \psi_1\ \until_I\ \psi_2 &&\text{iff } \exists\ t' \in t + I \text{ s.t. } w, t' \models \psi_2 \\
    & && && \phantom{iff} \text{ and } w, t'' \models \psi_1\ \forall\ t'' \in [t, t']
\end{alignat*}

A useful feature of STL is that, in addition to the Boolean semantics defined above, it also admits a \textit{quantitative semantics} giving the margin of satisfaction (or robustness margin) of an STL formula, denoted $\rho$. The formula is satisfied when $\rho > 0$ and not satisfied when $\rho < 0$. The robustness margin can also be defined inductively:
\begin{align*}
    \rho(\text{true}, s, t) &= \top \\
    \rho(\mu(x) \geq 0, s, t) &= \mu(s(t)) \\
    \rho(\neg\psi, s, t) &= -\rho(\psi, s, t) \\
    \rho(\psi_1 \wedge \psi_2, s, t) &= \min\{\rho(\psi_1, s, t), \rho(\psi_2, s, t)\} \\
    \rho(\psi_1 \until_I\ \psi_2, s, t) &= \sup_{t' \in t + I} \min\{\rho(\psi_2, s, t'), \inf_{t'' \in [t, t']} \rho(\psi_1, s, t'')
\end{align*}
where $\top$ is a constant taken to be greater than all other real values. In practice, linear-time algorithms exist for evaluating $\rho$ given a piecewise-affine signal $s$~\cite{donze13}.

It is important to make a distinction between the robustness margin of the specification, $\rho$, and the robustness of a plan designed to satisfy that specification. $\rho$ measures the margin by which the specification is met for a particular execution of a plan, but it does not provide much information about whether the specification will hold across multiple executions, particularly when external disturbances can affect those executions. In the next section, we formalize the robust planning problem, which aims at finding a plan that will satisfy the STL specification even when affected by external disturbances.

STL syntax may appear opaque at first glance, but its myriad symbols belie the fact that it is often straightforward to translate an STL formula into easily-understood natural language. For example, $\eventually_{[10, 20]} ((\always_{[0, 5]}\ x \geq 0)\ \until\ y \leq 0)$ can be read as ``between 10--\SI{20}{s} from now, $x$ must be positive for \SI{5}{s} before $y$ becomes negative.'' We provide more examples of STL formulae for robotics problems in Section~\ref{experiments}.

\section{Problem Statement}

In this paper, we focus on the problem of robust planning from an STL specification, which we view as a sequential two-player zero-sum game between the planner and its environment. In the first step of this game (planning time), the planner has the opportunity to tune a set of \textit{design parameters} $\theta$, but in the second step (run-time) the environment can change a distinct set of \textit{exogenous parameters} $\chi$ to degrade the performance of the plan. Together, $\theta \in \Theta$ and $\chi \in \mathcal{X}$ define the behavior of an autonomous system $\xi: \Theta \times \mathcal{X} \mapsto X^T$, which we assume is a known simulator function mapping design and exogenous parameters to a length-$T$ trace of states $x_t \in X$. We assume that $\xi$ is deterministic, so all uncertainty must be imported via $\chi$, but we assume that $\chi$ may be chosen \textit{adversarially} to degrade the performance of our chosen $\theta$ as much as possible. We also assume that the designer must commit to a choice of $\theta$ before the adversary chooses $\chi$ 

The performance of a plan is given by a cost function $J: X^T \mapsto \R$ assigning a scalar cost to a behavior trace. To accommodate STL specifications, we deal mainly with cost functions of the form
\begin{align*}
    J_\psi(\theta, \chi) = -\rho\pn{\psi, \xi(\theta, \chi)} + \lambda J_{other}(\theta, \chi)
\end{align*}
where $\rho\pn{\psi, \xi(\theta, \chi)}$ is the robustness margin of the behavior trace with respect to a given STL specification $\psi$. We negate $\rho$ so that minimizing $J$ maximizes the robustness margin, and the $\lambda J_{other}$ term permits us to consider other factors in the plan's performance (e.g. fuel use). The scaling factor $\lambda$ is typically small to prioritize satisfying the STL specification.

Since we assume that $\chi$ can vary adversarially to impose worst-case performance for any plan $\theta$, our goal is to find $\theta$ that are robust to this variation. Concretely, our goal is to solve an optimization problem representing a two-step sequential zero-sum game with two players:
\begin{align}
    \max_{\chi \in \mathcal{X}}\ \min_{\theta \in \Theta}\ J(\theta, \chi) \label{planning_problem}
\end{align}

To make this discussion concrete, consider a simple example of path planning for an aerial robot. In this case, $\psi$ might specify that we eventually ($\eventually$) reach a goal and always ($\always$) avoid some obstacles, $\theta$ might represent the locations of waypoints along the path and the parameters of a trajectory-tracking controller to follow those waypoints, and $\chi$ might represent the force from wind that attempts to drive the robot off course. The behavior $\xi$ might be a function that simulates the dynamics of the robot flying through wind, and the additional cost $J_{other}$ might impose a small penalty on large control inputs to conserve battery life. We provide more in-depth examples in Section~\ref{experiments}.

Our formulation differs from that presented in~\cite{Pant2018} and~\cite{Pantazides2022}; although both of these works seek to maximize the robustness margin $\rho$, neither consider the effect of disturbances $\chi$. Our formulation is also distinct from the mixed-integer formulation in~\cite{Sun2022}, since we consider $\rho$ as part of an objective rather than as a constraint. Our unconstrained approach does not provide the same completeness guarantees as a mixed-integer constrained optimization (used in~\cite{raman15,sadraddini15,Sun2022}), but empirical results in Section~\ref{experiments} demonstrate that our approach scales much better.

Of course, solving~\eqref{planning_problem} to global optimality in the general nonlinear case is intractable. Instead, we take advantage of this game structure to design an iterative algorithm to find the \textit{generalized Nash equilibrium}: the design parameters $\theta$ and corresponding $\chi$ such that neither the planner nor the adversary have an incentive to change their choice~\cite{Facchinei2007}. The next section describes this iterative algorithm, which we implement using nonlinear programming with differentiable simulation and differentiable temporal logic.

\section{Approach}

To solve the robust STL planning problem~\eqref{planning_problem}, we need to address two key points. First, we must develop a meta-heuristic to find a generalized Nash equilibrium of the sequential game~\eqref{planning_problem}, taking care that we do not overfit to any particular value of $\chi$.
We solve this challenge by developing an iterative counterexample-guided nonlinear optimization framework. Second, in order to solve this problem using nonlinear optimization, we need an efficient way to compute gradients of $J$ with respect to both $\theta$ and $\chi$, which requires us to differentiate not only the behavior function $\xi$ but also the robustness margin computation $\rho$. We address this challenge using differentiable programming, which we discuss next before introducing our high-level counterexample-guided optimization strategy.

\subsection{Differentiable Simulation and Temporal Logic}\label{autodiff}

Although it is possible to solve nonlinear optimization problems without access to the gradients of the objective or constraint functions, either by estimating gradients~\cite{suh2021_bundled_gradients} or using zero-order methods~\cite{nevergrad}, it is often much faster to use exact gradient information when it is available. However, exact gradients can be difficult to derive symbolically for complex optimization problems. Instead, recent works have turned to \textit{automatic differentiation} using differentiable programming to automatically compute gradients in problems such as 3D shape optimization~\cite{cascaval2021differentiable}, aircraft design optimization~\cite{sharpe_thesis}, robot design optimization~\cite{dawson2022architect1,du2021underwater}, and machine learning~\cite{jax2018github}.

Inspired by this trend, we implement $\xi$ using the JAX framework for automatic differentiation~\cite{jax2018github}, yielding a differentiable simulation of the underlying autonomous system. For a system where the behavior is defined by continuous-time dynamics $\dot{x} = f(x, \theta, \chi, t)$, implementing numerical integration in a differentiable language such as JAX allows us to automatically back-propagate through the simulator to find the gradients $\nabla_\theta \xi$ and $\nabla_\chi \xi$. These gradients can typically be computed much more quickly using automatic differentiation than by finite-difference methods~\cite{dawson2022architect1}.

We can use a similar differentiable programming approach to obtain gradients through the quantitative semantics of an STL specification. Before doing so, we must replace the discontinuous $\max$ and $\min$ operators used to compute $\rho$ with smooth approximations:
\begin{align*}
    \widetilde{\max}(x_1, x_2, \ldots) &= \frac{1}{k}\log\pn{e^{kx_1} + e^{kx_2} + \ldots} \\
    \widetilde{\min}(x_1, x_2, \ldots) &= -\widetilde{\max}(-x_1, -x_2, \ldots)
\end{align*}
where $k$ is a smoothing parameter and $\lim_{k\to\infty} \widetilde{\max} = \max$. This differentiable relaxation was introduced in~\cite{Pant2017} and later used in~\cite{Pant2018,Pantazides2022}; \cite{Leung2020} uses a slightly different approximation.

Using these smooth approximations, we implement the fast, linear-time algorithms for computing the robustness margin proposed by~\cite{donze13}, using the JAX framework to enable efficient automatic differentiation. In contrast to~\cite{Leung2020}, our method achieves computational complexity that is linear in the length of the state trace $T$ (the complexity of the \texttt{stlcg} framework in \cite{Leung2020} is quadratic in $T$ for the $\mathcal{U}$ operator).

By combining smooth approximations of STL quantitative semantics with differentiable programming, we can efficiently compute the gradients $\nabla_\theta \rho$ and $\nabla_\xi \rho$. By combining these gradients with those found using differentiable simulation, we can efficiently compute the gradient of the objective $J$ with respect to both the design parameters $\theta$ and the adversary's response $\chi$. Usefully, our use of differentiable programming means that we are not restricted to considering trajectory planning separately from the design of a tracking controller, as in~\cite{Pant2018} and~\cite{Sun2022}. Instead, we can consider an end-to-end gradient that combines the planned trajectory and controller parameters in $\theta$ and optimizes them jointly (see Section~\ref{experiments} for an example of this end-to-end optimization). In the next section, we discuss how end-to-end gradients enable an iterative algorithm for counterexample-guided robust optimization.

\subsection{Counterexample-guided Optimization}

To solve the planning problem in~\eqref{planning_problem}, we need to find a generalized Nash equilibrium between the planner and the adversary; i.e. values of $\theta$ and $\chi$ where neither we nor the adversary has any local incentive to change. A common solution strategy for such problems is the family of nonlinear Gauss-Seidel-type methods~\cite{Facchinei2007}. These methods solve max-min problems like~\eqref{planning_problem} by alternating between $\theta$ and $\chi$, tuning one set of parameters while keeping the others constant; i.e. alternating between the two optimization problems:
\begin{subequations}
    \begin{align}\label{eq:gauss_seidel}
        \theta^* &= \argmin_\theta J(\theta, \chi^*) \\
        \chi^* &= \argmax_\chi J(\theta^*, \chi)
    \end{align}
\end{subequations}
Although these methods are not guaranteed to converge, it is known that if they do, then the convergence point $(\theta^*, \chi^*)$ is a Nash equilibrium~\cite{Facchinei2007}.

A risk of applying such a simple alternating scheme is that the nonlinear optimization for both $\theta$ and $\chi$ can easily get caught in local minima. Such local minima not only reduce the performance of the optimized plan, but also increase the risk of ``overfitting'' to a particular value of $\chi^*$. This risk is particularly salient because the planner must commit to a choice of $\theta$ before the adversary has a final opportunity to choose $\chi$. To mitigate this risk and improve the robustness of our optimized plan, we extend a standard Gauss-Seidel method with two ideas from the machine learning and optimization literature. First, we take inspiration from the success of domain randomization in robust machine learning~\cite{tobin2017}: instead of optimizing $\theta$ with respect to a single fixed $\chi^*$, we can maintain a dataset $\mathcal{X}_N = \set{\chi_i}_{i=1,\ldots,N}$ and optimize the performance of $\theta$ across all of these samples:
\begin{subequations}
    \begin{align}\label{eq:gauss_seidel_domain_randomization}
        \theta^* &= \argmin_\theta \mathbb{E}_{\mathcal{X}_N} \left[ J(\theta, \chi_i) \right]\\
        \chi^* &= \argmax_\chi J(\theta^*, \chi)
    \end{align}
\end{subequations}

Incorporating domain randomization into the Gauss-Seidel method has the potential to improve the robustness of the resulting equilibria, but it is relatively sample inefficient; it may require a large number of random samples $\chi_i$. To address this sample inefficiency, we take inspiration from a second idea in the optimization and learning literature: learning from counterexamples~\cite{Chang2019}. The key insight here is that we can do better than simply randomly sampling $\chi_i$; we can use the values of $\chi^*$ found during successive iterations of the Gauss-Seidel process as high-quality counterexamples to guide the optimization of $\theta$. This insight results in our counterexample-guided Gauss-Seidel optimization method, which is outlined in pseudocode in Algorithm~\ref{alg:cg_gs}.

Our algorithm proceeds as follows. We begin by initializing the dataset with $N_0$ i.i.d. examples $\chi_i$, then we alternate between solving the two optimization problems in~\eqref{eq:gauss_seidel_domain_randomization}. At each iteration, we add our current estimate of the adversary's best response $\chi^*$ to the dataset, and we stop either when the algorithm reaches a fixed point (the adversary's best response after solving~\eqref{eq:gauss_seidel_domain_randomization} is the same as the best response from the previous round) or when a maximum number of iterations is reached. As we show experimentally in Section~\ref{experiments}, this counterexample-guided optimization achieves a higher sample efficiency than simple domain randomization, in that it finds plans that are more robust to adversarial disturbance while considering a much smaller dataset. Although our use of nonlinear optimization means that our algorithm is not complete, we find empirically that it succeeds in finding a satisfactory plan in the large majority of cases.

It is important to note that this algorithm is fundamentally enabled by the automatic differentiation approach detailed in Section~\ref{autodiff}; without access to the gradients of $J$ it would be much more difficult to solve the subproblems in lines~\ref{alg:opt_theta} and~\ref{alg:opt_chi} of Algorithm~\ref{alg:cg_gs}. Although some previous approaches obtain gradients of STL satisfaction with respect to $\theta$ using standard trajectory optimization formulations, as in~\cite{Pant2018}, we are not aware of any approaches that make use of gradients with respect to disturbance parameters. There has been some work on using counterexamples to guide mixed-integer planning~\cite{raman15}, but our experiments in the next section demonstrate that these mixed-integer programs are intractable for long horizon problems. Specifically, we find that solving even a single mixed-integer program can take more than an hour, so solving multiple programs to derive counterexamples is not a practical solution. In the next section, we demonstrate that our gradient-based counterexample-guided approach outperforms these existing approaches, not only finding more robust plans but requiring substantially less computation time to do so.

\begin{algorithm}
\caption{Counterexample-guided Gauss-Seidel method for solving robust planning problems}\label{alg:cg_gs}
\DontPrintSemicolon
    \KwInput{Starting dataset size $N_0$\\\phantom{Input: } Maximum number of iterations $M$}
    \KwOutput{Optimized design parameters $\theta^*$\\\phantom{Output: } Dataset of counterexamples $\mathcal{X}_N$}
    $\mathcal{X}_N \gets $ $N_0$ examples $\chi_i \in \mathcal{X}$ sampled uniformly i.i.d.\;
    $\chi^*_{prev} \gets \varnothing$\;
    \For{$i \in \set{1, \ldots, M}$}    
    { 
    	$\theta^* = \argmin_\theta \mathbb{E}_{\mathcal{X}_N} \left[ J(\theta, \chi_i) \right] \label{alg:opt_theta}$ \;
    	$\chi^* = \argmax_\chi J(\theta^*, \chi)$ \label{alg:opt_chi} \;
    	\If{$\chi^* = \chi^*_{prev}$}{\Break}
        $\chi^*_{prev} \gets \chi^*$\;
        Append $\chi^*$ to $\mathcal{X}_N$\;
    }
    \KwRet{$\theta^*$, $\mathcal{X}_N$}
\end{algorithm}

\section{Experiments}\label{experiments}

We validate our approach by means of two case studies involving the satellite rendezvous problem posed in~\cite{Jewison2016}. We benchmark against state-of-the-art planning algorithms to show the robustness and scalability benefits of our approach.

In this satellite rendezvous problem, the goal is to maneuver a chaser satellite to catch a target satellite. We can express this problem in the Clohessy-Wiltshire-Hill coordinate frame~\cite{Jewison2016}, which assumes that the target's orbit is circular and constructs a coordinate frame with the origin at the target, the $x$-axis pointing away from the Earth, the $y$-axis pointing along the target's orbit, and the $z$-axis pointing out of the orbital plane. In this frame, the chaser's dynamics are approximately linear, with positions $p_x$, $p_y$, $p_z$ and velocities $v_x$, $v_y$, $v_z$ varying according to controlled thrust in each direction $u_x$, $u_y$, $u_z$:
\begin{align*}
    \mat{\dot{p}_x \\ \dot{p}_y \\ \dot{p}_z \\ \dot{v}_x \\ \dot{v}_y \\ \dot{v}_z} = \mat{
        v_x \\
        v_y \\
        v_z \\
        3n^2 p_x + 2n v_y + u_x / m\\
        -2n v_x + u_y / m\\
        -n^2 p_z + u_z / m
    }
\end{align*}
$n = \sqrt{\mu / a^3}$ is the mean-motion of the target, determined by the Earth's gravitational constant $\mu = \SI{3.986e14}{m^3/s^2}$ and the target's altitude $a$ (i.e. the length of the semi-major orbital axis, \SI{353}{km} in low Earth orbit). $m = \SI{500}{kg}$ is the mass of the chaser satellite~\cite{Jewison2016}.

\begin{figure}[t]
    \centering
    \includegraphics[width=\linewidth]{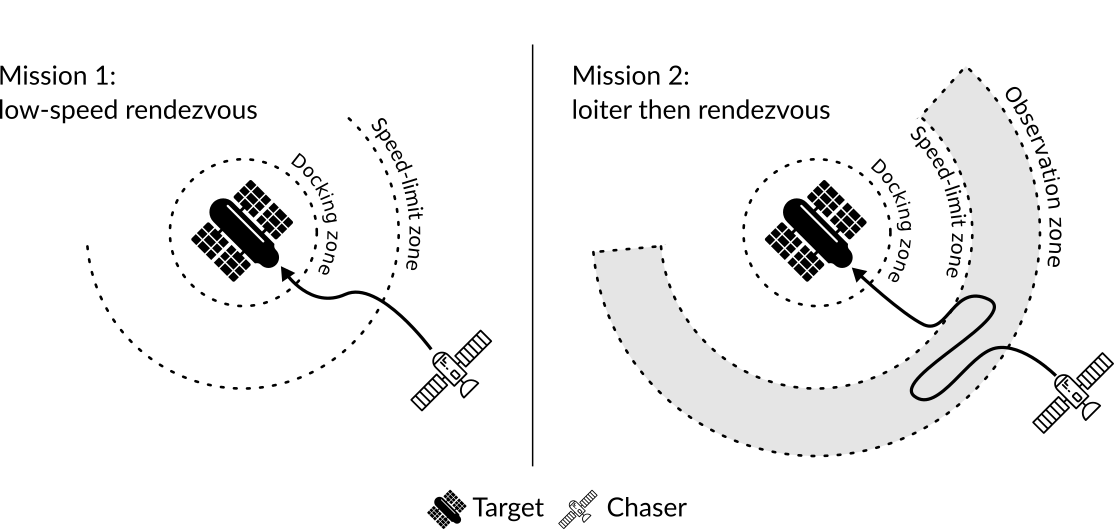}
    \caption{Two satellite rendezvous missions used to test our framework. In the first mission, the chaser satellite must eventually reach the target while respecting a maximum speed constraint in the region immediately around the target. In the second mission, the chaser must still reach the target and obey the speed limit, but it must also loiter in an observation region for some minimum time before approaching. The first mission requires an STL formula with three predicates and three temporal operators, while the second mission requires five predicates and five temporal operators.}
    \label{fig:mission_specs}
\end{figure}

In this setting, we construct STL specifications for two rendezvous missions: a simple low-speed rendezvous and a more complex loiter-then-rendezvous mission, illustrated in Fig.~\ref{fig:mission_specs}. The STL specifications for each mission, $\psi_1$ and $\psi_2$, are given formally as:
\begin{align*}
    \psi_1 &= \psi_\text{reach target} \wedge \psi_\text{speed limit} \\
    \psi_2 &= \psi_\text{reach target} \wedge \psi_\text{speed limit} \wedge \psi_\text{loiter} \\
    \psi_\text{reach target} &= \eventually \pn{r \leq 0.1} \\
    \psi_\text{speed limit} &= \pn{r \geq 2.0} \until\ \always \pn{v \leq 0.1} \\
    \psi_\text{loiter} &= \eventually \always_{[0, T_{obs}]} \pn{2.0 \leq r \wedge r \leq 3.0}
\end{align*}
where $r = \sqrt{p_x^2 + p_y^2 + p_z^2}$ and $v = \sqrt{v_x^2 + v_y^2 + v_z^2}$. Informally, $\psi_1$ requires that ``the chaser eventually comes within \SI{0.1}{m} of the target and does not come within \SI{2}{m} of the target until its speed is less than \SI{0.1}{m/s}'', and $\psi_2$ additionally requires that ``the chaser eventually spends at least $T_{obs}$ seconds in the region between 2--\SI{3}{m} from the target.'' For each mission, the design parameters $\theta$ include both state/input waypoints along a planned trajectory and the feedback gains used to track that trajectory, and the exogenous parameters $\chi$ represent bounded uncertainty in the initial state of the chaser ($p_x(0), p_y(0) \in [10, 13]$, $p_z(0) \in [-3, 3]$, $v_x(0), v_y(0), v_z(0) \in [-1, 1]$). We use a \SI{200}{s}-long simulation with a \SI{2}{s} timestep for both missions, and $T_{obs} = \SI{10}{s}$.

\begin{figure}[t]
    \centering
    \includegraphics[width=0.9\linewidth]{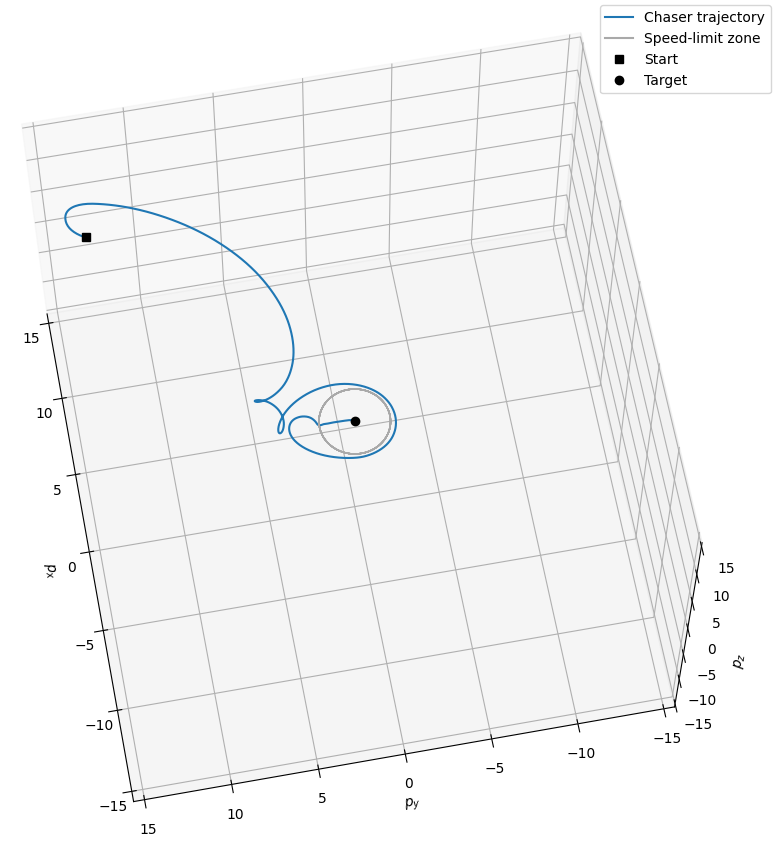}
    \caption{The optimized trajectory found using our counterexample-guided optimization strategy for mission 1 (rendezvous with speed constraint). The chaser satellite only enters the speed-limit zone once it has slowed down sufficiently.}
    \label{fig:mission_1_traj}
\end{figure}

\begin{figure}[t]
    \centering
    \includegraphics[width=\linewidth]{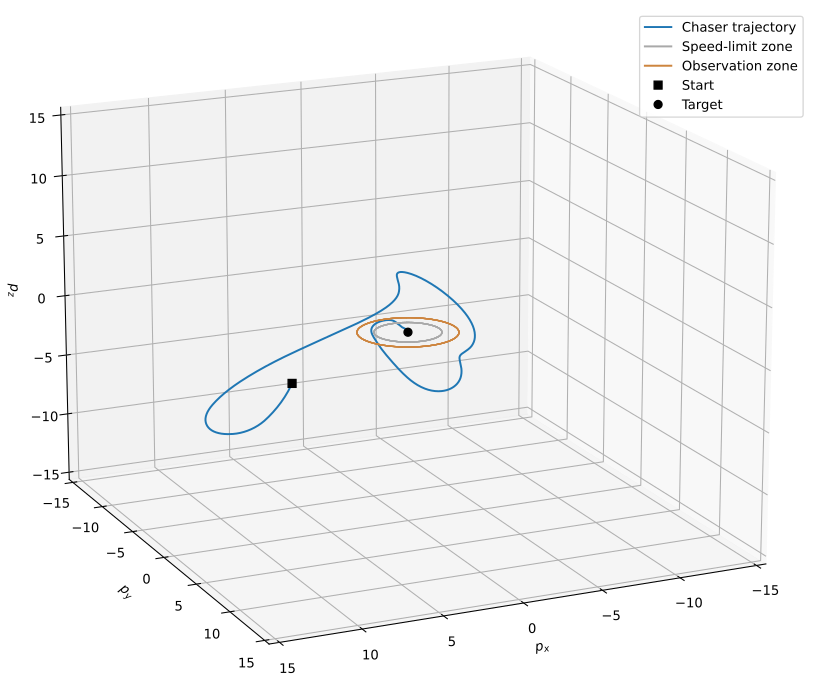}
    \caption{The optimized trajectory found using our counterexample-guided optimization strategy for mission 2 (loiter then rendezvous with speed constraint). The optimized plan satisfies the additional mission requirement of spending time in the observation region before approaching the target.}
    \label{fig:mission_2_traj}
\end{figure}

For each mission $i=1, 2$, we define a cost function as $J_i = \rho_i + \lambda I$, where $\rho_i = \rho(\psi_i, \xi(\theta, \chi), 0)$ is the STL robustness margin at the start of the trajectory, $I$ is the total impulse required to execute the maneuver (in Newton-seconds), and $\lambda = 5\times10^{-5}$. By applying our iterative counterexample-guided optimization strategy to this problem, we find the optimized trajectories for mission 1 and 2 shown in Figs.~\ref{fig:mission_1_traj} and~\ref{fig:mission_2_traj} along with the worst-case $\chi$. In these examples, we use $N_0=8$ initial examples and $M=10$ maximum rounds, but the algorithm converges in less than 10 rounds in all trials. In both missions, our approach reliably finds a solution that remains feasible despite worst-case variation in the exogenous parameters, achieving a positive STL robustness margin in $>90\%$ of trials in each case. Our counterexample-guided approach requires an average of \SI{53.7}{s} to solve mission 1 and \SI{194.2}{s} to solve mission 2 (averaged across 50 trials).

\begin{figure}[t]
    \centering
    \includegraphics[width=\linewidth]{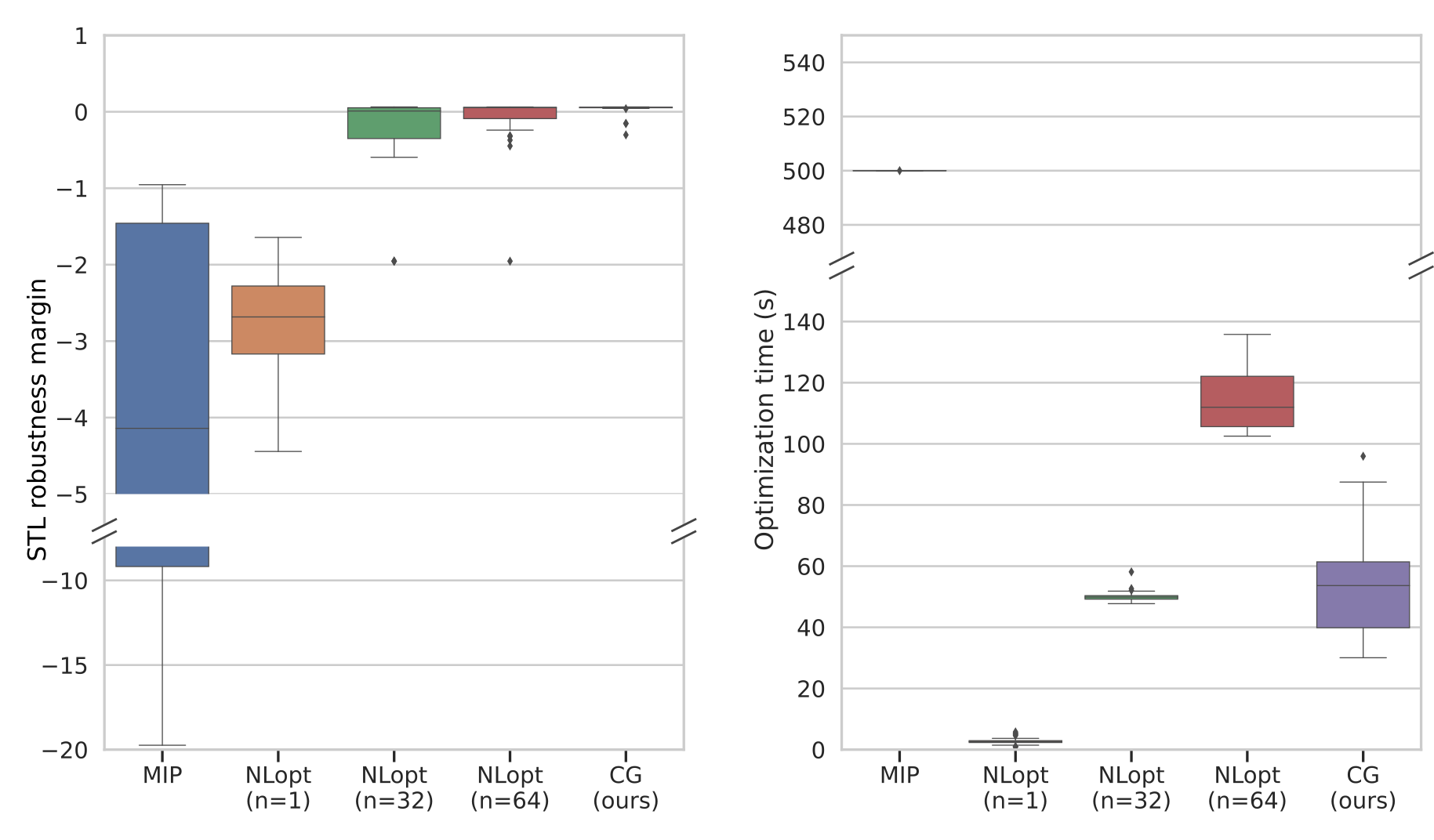}
    \caption{Comparison of different STL planning methods on the first example mission, averaged over 50 random seeds. Left: the robustness margin $\rho(\psi_1)$ computed for the optimized design parameters and worst-case exogenous parameters. Right: the planning time required by each method. Our method (CG) achieves much higher robustness than all other methods (satisfying the STL specification despite adversarial perturbations in all but 3 instances) and runs twice as fast as the next-most-robust method.}
    \label{fig:mission_1_comparison}
\end{figure}

We can quantitatively compare our approach against two state of the art approaches: a mixed-integer STL planner based on that in~\cite{raman15} and~\cite{sadraddini15} and the nonlinear optimization approach in~\cite{Pant2018,Pantazides2022}. The mixed-integer planner (MIP) in~\cite{raman15} uses a model-predictive control formulation and proposes to add counterexamples after solving each instance of the mixed-integer program; however, we found that even a single instance could not be solved to optimality within 1 hour for either mission, and so we compare with the best solution found within a given period of time. Even though the size of the mixed-integer program in~\cite{raman15} grows linearly with the horizon of the problem, the complexity of solving the resulting MIP grows exponentially in the number of integer variables (these problems require between 2800--4500 integer variables when solved using Gurobi). Since it was not tractable to solve even once instance of the MIP, we were unable to use any MIP-generated counterexamples as proposed in~\cite{raman15}; instead, we take the best feasible solution found after \SI{500}{s} for the first mission and after \SI{1000}{s} for the second mission.

We also compare with an extension of the nonlinear optimization from~\cite{Pant2018,Pantazides2022}, where we add domain randomization to the authors' existing trajectory optimization formulation, averaging the objective over either 32 or 64 different values of $\chi$. We note that these methods rely on a similar optimization approach as those proposed in~\cite{Leung2020}, but we re-implement the authors' method to ensure a fair comparison (our implementation makes use of just-in-time compilation to speed the optimization process, and so comparing with off-the-shelf implementations like \texttt{stlcg}~\cite{Leung2020} would not be fair).

\begin{figure}[t]
    \centering
    \includegraphics[width=\linewidth]{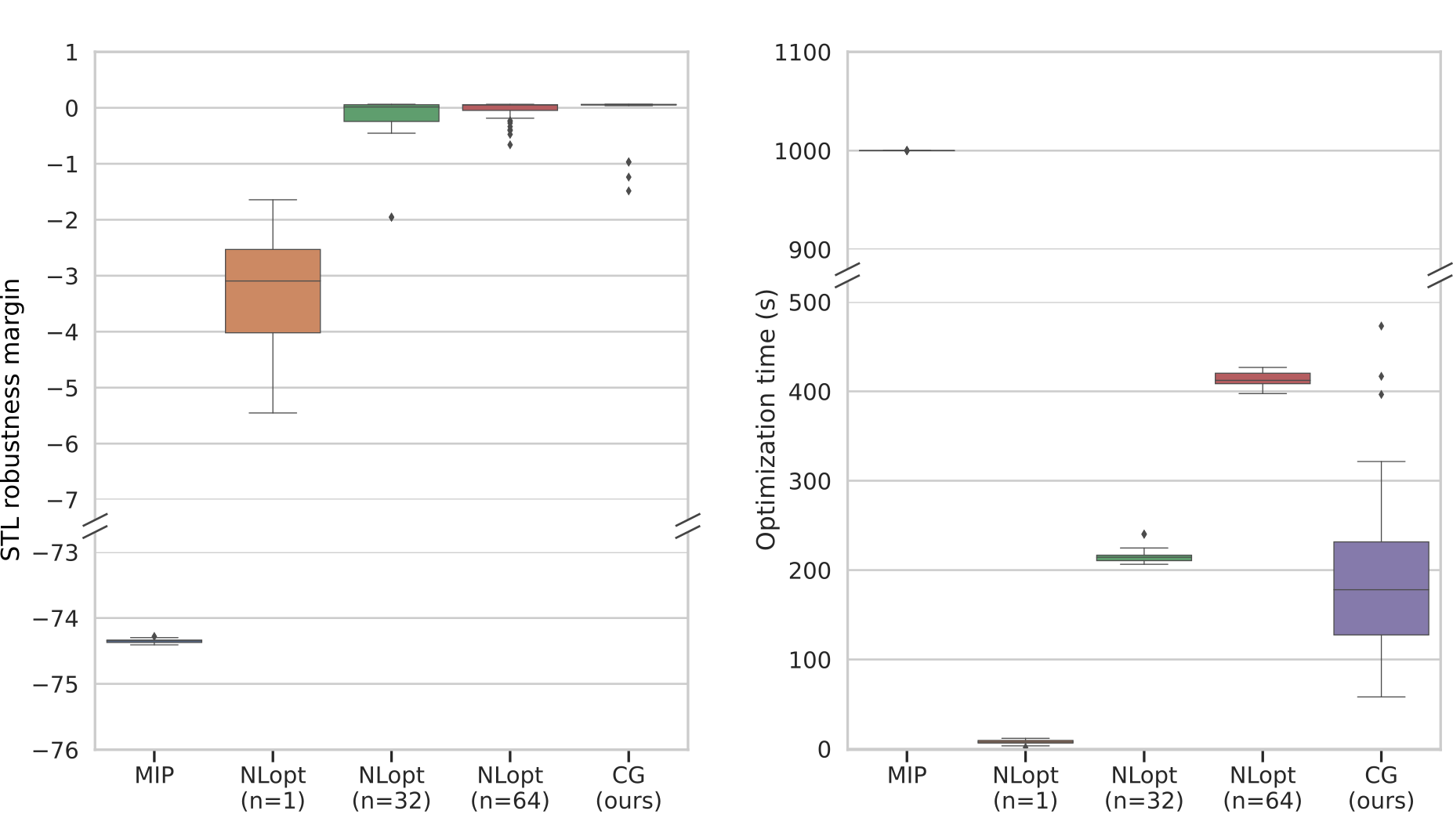}
    \caption{Comparison of different STL planning methods on the second example mission, averaged over 50 random seeds. Left: the robustness margin $\rho(\psi_2)$ computed for the optimized design parameters and worst-case exogenous parameters. Right: time required by each method to find a plan. Our method (CG) finds much more robust plans, satisfying the specification in all but 4 instances compared to 17 failures for the next-best method (NLopt with 64 examples). Our method also runs more than twice as fast as the next-most-robust method.}
    \label{fig:mission_2_comparison}
\end{figure}

A comparison of our counterexample-guided approach (CG), nonlinear optimization with domain randomization (NLopt), and the mixed-integer planner (MIP) is shown in Fig.~\ref{fig:mission_1_comparison} for the first mission and Fig.~\ref{fig:mission_2_comparison} for the second mission. In all cases, we compare across 50 random trials, computing the time required to solve each instance and the robustness of the optimized plan when subject to adversarial disturbances. All experiments were run on a laptop computer with \SI{8}{GB} RAM and a \SI{1.8}{GHz} 8-core processor.

We find that our method is consistently more robust than prior methods; in the first mission, it satisfies the STL specification in all but 3 trials, despite adversarial disturbances. For comparison, the next-best method (NLopt with domain randomization across 64 examples) failed to solve the first mission in 14 out of 50 trials and took more than twice as long on average to find a plan (\SI{114.3}{s} as opposed to \SI{53.7}{s} for our method). This advantage is due to the quality of the examples used during optimization; instead of 64 random samples, our method uses 8 initial random samples and between 1 and 4 counterexamples (median 2) representing worst-case variation in $\chi$, making our method much more sample-efficient.

We also find that our method finds more robust solutions than the MIP method, since MIP cannot tractably consider variation in $\chi$ (the MIP method is also unable to find a feasible solution within \SI{500}{s} in 16 out of 50 trials). MIP's performance also suffers due to discretization error, since we were forced to discretize the continuous-time dynamics with relatively few knot points (one every \SI{2}{s}) to yield a tractable MIP optimization problem.

Our method also performs well on the second mission planning problem: only our method consistently finds solutions that are robust to variation in $\chi$ (see Fig.~\ref{fig:mission_2_comparison}). Due to the increased complexity of this example, the MIP method finds a feasible solution within \SI{1000}{s} in only 16 out of 50 trials (the MIP encoding of this mission requires 7769 continuous variables, 2806 binary variables, and 1479 constraints), and the feasible solutions found within \SI{1000}{s} tend to be of poor quality. The second-most-robust method, NLopt with 64 random examples, takes more than twice as long as our method and fails to satisfy the STL specification in 17 out of 50 trials (compared to only 4 failures for our method). Our method required a median of 2 counterexamples in addition to the 8 initial examples to solve this planning problem (the slowest trial required 7 additional examples).

These data demonstrate that our counterexample-guided approach to planning from STL specifications is faster, more sample efficient, and more robust to adversarial disturbances than state-of-the-art approaches. A software implementation of our method is available online at \url{https://github.com/MIT-REALM/architect}.

\subsection{Hardware Demonstration}

We also validate our approach in hardware by solving the loiter-then-rendezvous mission for a Turtlebot 3 ground robot. We replace the satellite dynamics with nonlinear Dubins dynamics and plan a trajectory that satisfies $\psi = \psi_\text{reach target} \wedge \psi_\text{loiter}$ (we do not include the speed limit because the Turtlebot already has a relatively small maximum speed). Our counterexample-guided planner solves this problem in \SI{26.22}{s}, yielding the trajectory shown in Fig.~\ref{fig:hw}. A video of this demonstration is included in the supplementary materials.

\begin{figure}[t]
    \centering
    \includegraphics[width=\linewidth]{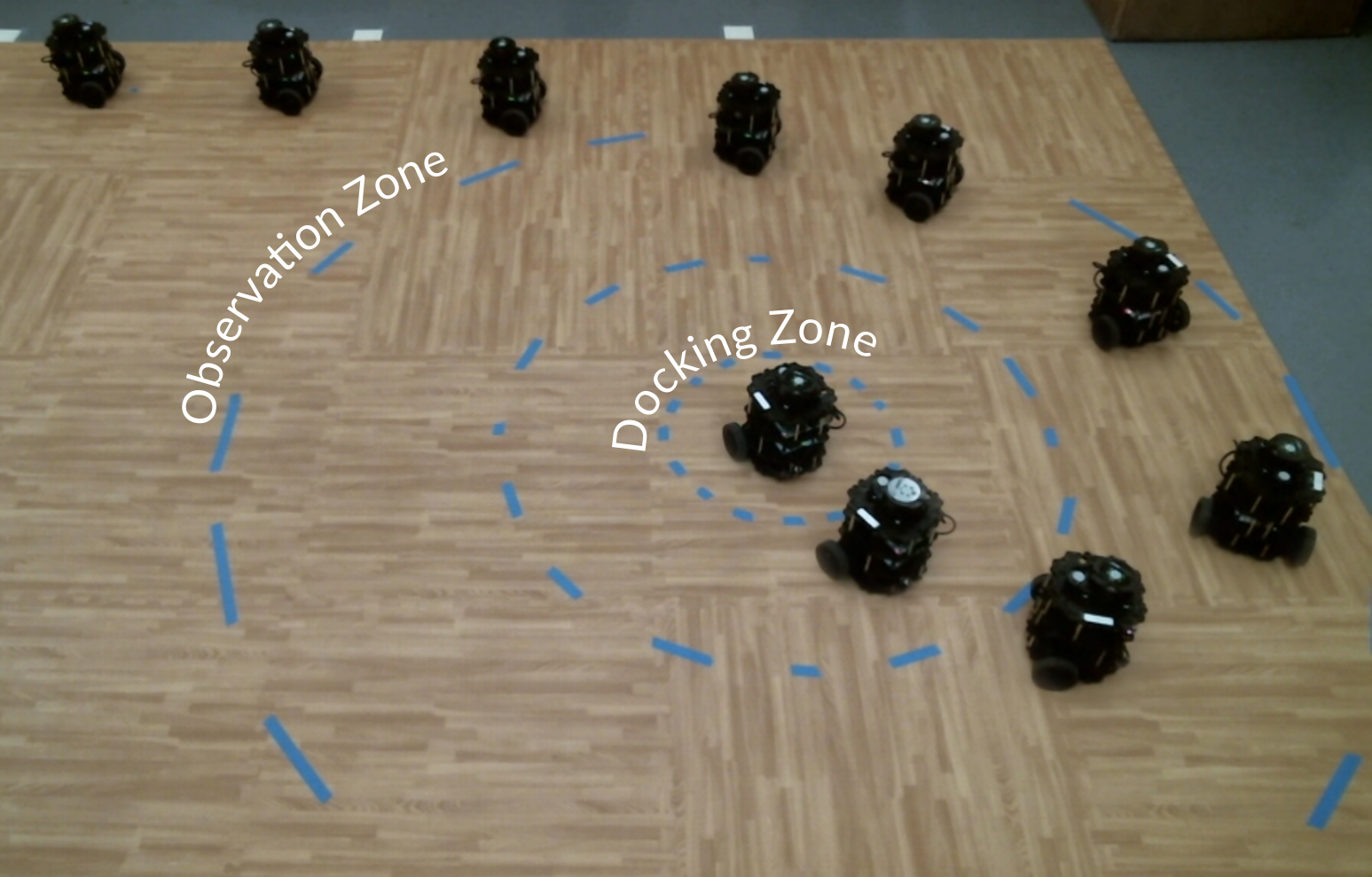}
    \caption{The trajectory found using our counterexample-guided planner successfully moves the robot through the observation zone (where it must spend at least \SI{10}{s}) and into the docking zone. Odometry data indicate that the planned trajectory achieves an STL robustness margin of $0.0255$.}
    \label{fig:hw}
\end{figure}

\section{Conclusion}

In this paper, we introduce a novel framework for robust optimization-based planning from temporal logic specifications. We frame the robust planning problem as a sequential two-player game between the planner, which chooses a set of design parameters at planning time, and the environment, which picks disturbances adversarially in response to the planner's choice. We develop an iterative counterexample-guided algorithm to find plans that robustly satisfy temporal logic specifications despite worst-case disturbances from the environment. Our method, which relies on differentiable programming for simulation and evaluating the temporal logic specification, not only finds more robust plans than state-of-the-art methods but also runs substantially faster. We apply our method to two planning problems involving time horizons $>$\SI{100}{s} and STL specifications with multiple nested temporal operators, and we provide source code for our approach at \url{https://github.com/MIT-REALM/architect}.

In future work, we hope to extend our framework to offer completeness guarantees by combining our counterexample-guided optimization with complete model checking and STL falsification methods, such as the Breach framework~\cite{donze10}. We also look forward to exploring applications of this framework to problems involving multiple agents, particularly for human-robot interaction, and to combining local gradient-based optimization with sampling based methods to solve more complex task-and-motion planning problems.

\bibliographystyle{IEEEtran}
\bibliography{main}

\begin{thebibliography}{10}
\providecommand{\url}[1]{#1}
\csname url@samestyle\endcsname
\providecommand{\newblock}{\relax}
\providecommand{\bibinfo}[2]{#2}
\providecommand{\BIBentrySTDinterwordspacing}{\spaceskip=0pt\relax}
\providecommand{\BIBentryALTinterwordstretchfactor}{4}
\providecommand{\BIBentryALTinterwordspacing}{\spaceskip=\fontdimen2\font plus
\BIBentryALTinterwordstretchfactor\fontdimen3\font minus
  \fontdimen4\font\relax}
\providecommand{\BIBforeignlanguage}[2]{{%
\expandafter\ifx\csname l@#1\endcsname\relax
\typeout{** WARNING: IEEEtran.bst: No hyphenation pattern has been}%
\typeout{** loaded for the language `#1'. Using the pattern for}%
\typeout{** the default language instead.}%
\else
\language=\csname l@#1\endcsname
\fi
#2}}
\providecommand{\BIBdecl}{\relax}
\BIBdecl

\bibitem{donze10}
\BIBentryALTinterwordspacing
A.~Donzé, ``Breach, a toolbox for verification and parameter synthesis of
  hybrid systems,'' \emph{Lecture Notes in Computer Science (including
  subseries Lecture Notes in Artificial Intelligence and Lecture Notes in
  Bioinformatics)}, vol. 6174 LNCS, pp. 167--170, 2010. [Online]. Available:
  \url{https://link-springer-com.libproxy.mit.edu/chapter/10.1007/978-3-642-14295-6_17}
\BIBentrySTDinterwordspacing

\bibitem{Sun2022}
\BIBentryALTinterwordspacing
D.~Sun, J.~Chen, S.~Mitra, and C.~Fan, ``Multi-agent motion planning from
  signal temporal logic specifications,'' \emph{IEEE Robotics and Automation
  Letters (RA-L)}, 1 2022. [Online]. Available:
  \url{https://arxiv.org/abs/2201.05247v1}
\BIBentrySTDinterwordspacing

\bibitem{Pant2017}
\BIBentryALTinterwordspacing
Y.~V. Pant, H.~Abbas, and R.~Mangharam, ``Smooth operator: Control using the
  smooth robustness of temporal logic,'' \emph{IEEE Conference on Control
  Technology and Applications, 2017}, 8 2017. [Online]. Available:
  \url{https://repository.upenn.edu/mlab\_papers/100}
\BIBentrySTDinterwordspacing

\bibitem{Pant2018}
\BIBentryALTinterwordspacing
Y.~V. Pant, H.~Abbas, R.~A. Quaye, and R.~Mangharam, ``Fly-by-logic: Control of
  multi-drone fleets with temporal logic objectives,'' \emph{Proceedings - 9th
  ACM/IEEE International Conference on Cyber-Physical Systems, ICCPS 2018}, pp.
  186--197, 8 2018. [Online]. Available:
  \url{https://phys.org/news/2016-12-traffic-}
\BIBentrySTDinterwordspacing

\bibitem{Pantazides2022}
\BIBentryALTinterwordspacing
A.~Pantazides, D.~Aksaray, and D.~Gebre-egziabher, \emph{Satellite Mission
  Planning with Signal Temporal Logic Specifications}.\hskip 1em plus 0.5em
  minus 0.4em\relax AIAA, 2022. [Online]. Available:
  \url{https://arc.aiaa.org/doi/abs/10.2514/6.2022-1091}
\BIBentrySTDinterwordspacing

\bibitem{Plaku2016}
E.~Plaku and S.~Karaman, ``Motion planning with temporal-logic specifications:
  Progress and challenges,'' \emph{AI Communications}, vol.~29, pp. 151--162, 1
  2016.

\bibitem{Takano2021}
R.~Takano, H.~Oyama, and M.~Yamakita, ``Continuous optimization-based task and
  motion planning with signal temporal logic specifications for sequential
  manipulation.''\hskip 1em plus 0.5em minus 0.4em\relax Institute of
  Electrical and Electronics Engineers (IEEE), 10 2021, pp. 8409--8415.

\bibitem{yang20}
G.~Yang, C.~Belta, and R.~Tron, ``Continuous-time signal temporal logic
  planning with control barrier functions,'' in \emph{2020 American Control
  Conference (ACC)}, 2020, pp. 4612--4618.

\bibitem{Leung2020}
\BIBentryALTinterwordspacing
K.~Leung, N.~Arechiga, and M.~Pavone, ``Backpropagation through signal temporal
  logic specifications: Infusing logical structure into gradient-based
  methods,'' \emph{Springer Proceedings in Advanced Robotics}, vol.~17, pp.
  432--449, 7 2020. [Online]. Available:
  \url{https://arxiv.org/abs/2008.00097v3}
\BIBentrySTDinterwordspacing

\bibitem{kantaros20}
\BIBentryALTinterwordspacing
Y.~Kantaros and M.~M. Zavlanos, ``Stylus*: A temporal logic optimal control
  synthesis algorithm for large-scale multi-robot systems,'' \emph{The
  International Journal of Robotics Research}, vol.~39, no.~7, pp. 812--836,
  2020. [Online]. Available: \url{https://doi.org/10.1177/0278364920913922}
\BIBentrySTDinterwordspacing

\bibitem{vasile17}
C.-I. Vasile, V.~Raman, and S.~Karaman, ``Sampling-based synthesis of
  maximally-satisfying controllers for temporal logic specifications,'' in
  \emph{2017 IEEE/RSJ International Conference on Intelligent Robots and
  Systems (IROS)}, 2017, pp. 3840--3847.

\bibitem{raman15}
\BIBentryALTinterwordspacing
V.~Raman, A.~Donz\'{e}, D.~Sadigh, R.~M. Murray, and S.~A. Seshia, ``Reactive
  synthesis from signal temporal logic specifications,'' in \emph{Proceedings
  of the 18th International Conference on Hybrid Systems: Computation and
  Control}, ser. HSCC '15.\hskip 1em plus 0.5em minus 0.4em\relax New York, NY,
  USA: Association for Computing Machinery, 2015, p. 239–248. [Online].
  Available: \url{https://doi.org/10.1145/2728606.2728628}
\BIBentrySTDinterwordspacing

\bibitem{sadraddini15}
S.~Sadraddini and C.~Belta, ``Robust temporal logic model predictive control,''
  in \emph{2015 53rd Annual Allerton Conference on Communication, Control, and
  Computing (Allerton)}, 2015, pp. 772--779.

\bibitem{donze13}
\BIBentryALTinterwordspacing
A.~Donzé, T.~Ferrère, and O.~Maler, ``Efficient robust monitoring for stl,''
  \emph{Lecture Notes in Computer Science (including subseries Lecture Notes in
  Artificial Intelligence and Lecture Notes in Bioinformatics)}, vol. 8044
  LNCS, pp. 264--279, 2013. [Online]. Available:
  \url{https://link.springer.com/chapter/10.1007/978-3-642-39799-8\_19}
\BIBentrySTDinterwordspacing

\bibitem{Facchinei2007}
\BIBentryALTinterwordspacing
F.~Facchinei and C.~Kanzow, ``Generalized nash equilibrium problems,''
  \emph{4OR 2007 5:3}, vol.~5, pp. 173--210, 9 2007. [Online]. Available:
  \url{https://link.springer.com/article/10.1007/s10288-007-0054-4}
\BIBentrySTDinterwordspacing

\bibitem{suh2021_bundled_gradients}
H.~Suh, T.~Pang, and R.~Tedrake, ``Bundled gradients through contact via
  randomized smoothing,'' \emph{ArXiv}, vol. abs/2109.05143, 2021.

\bibitem{nevergrad}
J.~Rapin and O.~Teytaud, ``{Nevergrad - A gradient-free optimization
  platform},'' \url{https://GitHub.com/FacebookResearch/Nevergrad}, 2018.

\bibitem{cascaval2021differentiable}
D.~Cascaval, M.~Shalah, P.~Quinn, R.~Bodik, M.~Agrawala, and A.~Schulz,
  ``Differentiable 3d cad programs for bidirectional editing,'' \emph{arXiv},
  vol. abs/2110.01182, 2021.

\bibitem{sharpe_thesis}
P.~D. Sharpe, ``Aerosandbox: A differentiable framework for aircraft design
  optimization,'' Master's thesis, MIT, 2021.

\bibitem{dawson2022architect1}
C.~Dawson and C.~Fan, ``Certifiable robot design optimization using
  differentiable programming,'' \emph{Under Review}, 2022.

\bibitem{du2021underwater}
T.~Du, J.~Hughes, S.~Wah, W.~Matusik, and D.~Rus, ``Underwater soft robot
  modeling and control with differentiable simulation,'' \emph{IEEE Robotics
  and Automation Letters}, 2021.

\bibitem{jax2018github}
\BIBentryALTinterwordspacing
J.~Bradbury, R.~Frostig, P.~Hawkins, M.~J. Johnson, C.~Leary, D.~Maclaurin,
  G.~Necula, A.~Paszke, J.~Vander{P}las, S.~Wanderman-{M}ilne, and Q.~Zhang,
  ``{JAX}: composable transformations of {P}ython+{N}um{P}y programs,'' 2018.
  [Online]. Available: \url{http://github.com/google/jax}
\BIBentrySTDinterwordspacing

\bibitem{tobin2017}
J.~Tobin, R.~Fong, A.~Ray, J.~Schneider, W.~Zaremba, and P.~Abbeel, ``Domain
  randomization for transferring deep neural networks from simulation to the
  real world,'' in \emph{2017 IEEE/RSJ International Conference on Intelligent
  Robots and Systems (IROS)}, 2017, pp. 23--30.

\bibitem{Chang2019}
\BIBentryALTinterwordspacing
Y.-C. Chang, N.~Roohi, and S.~Gao, ``{Neural Lyapunov Control},'' in
  \emph{Advances in Neural Information Processing Systems}, vol.~32, 2019, pp.
  3245--3254. [Online]. Available:
  \url{https://github.com/YaChienChang/Neural-Lyapunov-Control}
\BIBentrySTDinterwordspacing

\bibitem{Jewison2016}
C.~Jewison and R.~S. Erwin, ``A spacecraft benchmark problem for hybrid control
  and estimation,'' \emph{2016 IEEE 55th Conference on Decision and Control,
  CDC 2016}, pp. 3300--3305, 12 2016.

\end{thebibliography}

\end{document}